\documentclass[runningheads]{llncs}

\usepackage{eccv}

\usepackage{eccvabbrv}

\usepackage{graphicx}
\usepackage{booktabs}
\usepackage{multirow}
\usepackage[accsupp]{axessibility}  
\usepackage[table]{xcolor}

\usepackage{hyperref}
\usepackage{array}
\usepackage{orcidlink}
\usepackage{wrapfig}
\usepackage{diagbox} 
\usepackage{enumitem}

\begin{document}

\title{Towards High-Resolution Visual Perception via Hierarchical Entity Exploration} 

\titlerunning{Hierarchical Entity Exploration}

\author{Ziyu Ma\inst{1}$^{\star}$ \and
Shidong Yang\inst{1}$^{\star}$ \and
Yuxiang Ji\inst{1} \and
Yiming Hu\inst{1}$^{\dagger}$ \and
Tongwen Huang\inst{1}$^{\dagger}$ \and
Yong Wang\inst{1}$^{\ddagger}$ \and
Jianfei Cai\inst{2} \and
Xiangxiang Chu\inst{1}}

\authorrunning{Z.~Ma, S.~Yang et al.}

\institute{AMAP, Alibaba Group, China \and
Monash University, Australia}


\maketitle

\let\oldthefootnote\thefootnote
\let\thefootnote\relax
\footnotetext{$^{\star}$Equal contribution.\quad $^{\dagger}$Project lead. \quad $^{\ddagger}$Corresponding author.}
\let\thefootnote\oldthefootnote

\begin{abstract}
  High-resolution (HR) image perception remains a key challenge in multimodal large language models (MLLMs), as fine-grained details are often lost when the image is processed as a whole. Existing methods either require training to teach models where to look or heuristically divide the image into fixed regions, both of which struggle to generalize in complex HR scenes. In this work, we propose Hierarchical Entity Exploration (HEE), a training-free and model-agnostic framework that transforms static image understanding into dynamic, query-guided entity exploration. HEE first evaluates each region using a dual scoring mechanism to determine whether it already contains sufficient evidence to answer the question. If not, it applies object detection within the most promising region to extract fine-grained entities, clusters them into coherent subregions, and organizes them into a multi-level semantic hierarchy for deeper exploration. When deeper regions still fail to yield confident answers, a confidence-guided backtracking mechanism revisits alternative paths to ensure adaptive perception. Extensive results show that HEE outperforms training-free methods like ZoomEye and RAP in both accuracy and efficiency on two complex HR benchmarks (Visual Probe and HR-Bench), across different MLLMs such as Qwen2.5-VL and LLaVA-OneVision. Moreover, HEE demonstrates generalization on the MME-RealWorld benchmark.
\end{abstract}

\section{Introduction}
\label{sec:intro}
Current MLLMs~\cite{liu2024improved,Qwen-VL,liu2024llavanext,ma2026skillclaw,wang2023cogvlm,abdin2024phi,Ma_Gou_Hu_Wang_Zhuang_Cai_2026} operate under a fixed input resolution (e.g., $448\times448$), a design choice that simplifies computation but compromises the fidelity of high-resolution (HR) images. When HR inputs are uniformly resized, geometric deformation and fine-detail loss inevitably occur, leading to noticeable performance drops in tasks that rely on precise visual cues, such as visual grounding and OCR, where subtle structures are essential~\cite{zhang2024llava,ma2024drvideo,jing2023deep,10.1145/3664647.3681403,ma2024gerea}.

To enhance the model's perception of image details, existing methods often adopt a ``locate-then-zoom-in" strategy. Training-based paradigms (e.g., supervised fine-tuning~\cite{visual-cot} and reinforcement learning~\cite{deepeyes,unvisual_cot,ji2026tree}) explicitly teach models to focus on key regions by introducing visual tool-use abilities and corresponding training data. A representative method is DeepEyes~\cite{deepeyes}, which defines a zoom-in tool and optimizes its usage through end-to-end reinforcement learning. During inference, the model actively selects and enlarges task-relevant regions via the visual tool to progressively focus on fine-grained information in high-resolution images. However, such training-based paradigms suffer from critical drawbacks \cite{investigating,limit-of-rlvr}, including high costs, lengthy training, and a lack of cross-architecture transferability, which severely limits their practical deployment.

\begin{figure*}[t]
    \centering
    \includegraphics[scale=0.35]{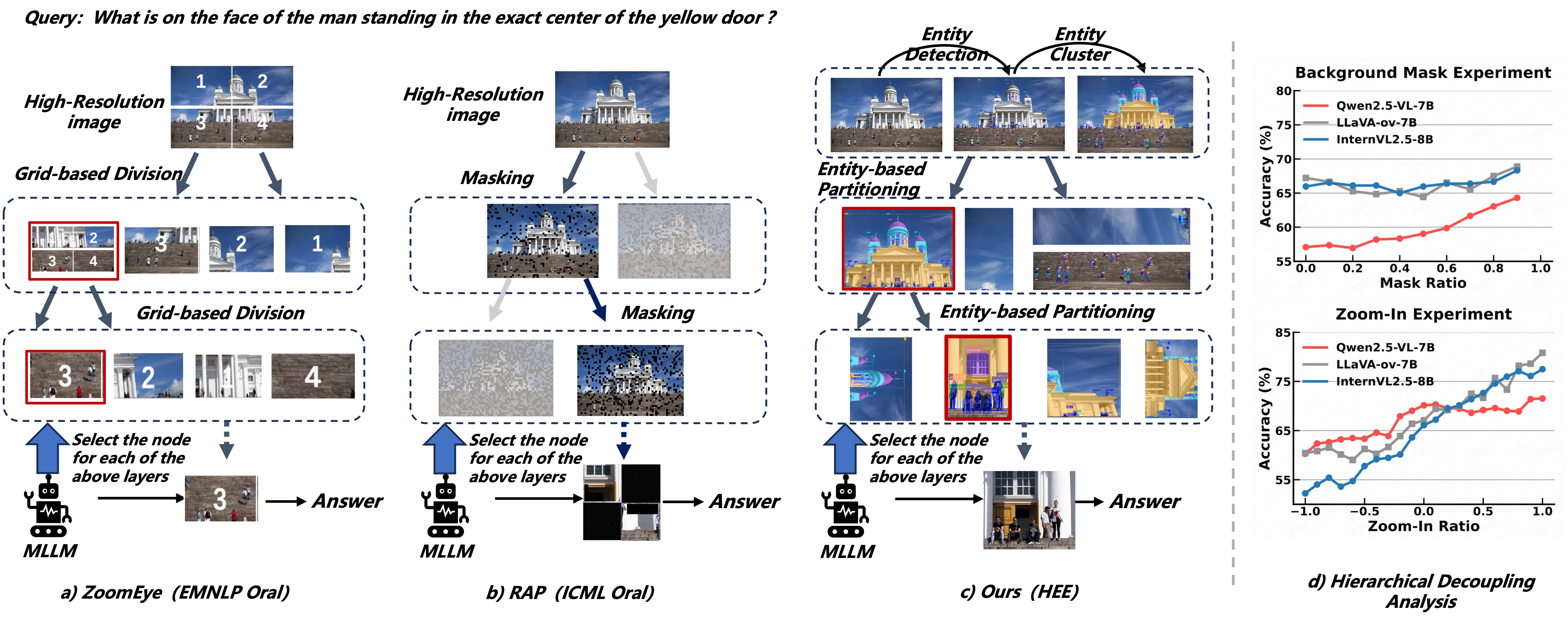}

    \caption{ Illustration of two common approaches (i.e., ZoomEye, and RAP) for handling high-resolution (HR) images, alongside our HEE. The right subfigures show our hierarchical decoupling analysis, including the background masking and zoom-in experiment.}
    \label{hee_fig1}
\vspace{-10pt}     
\end{figure*}

Apart from studies that enhance perception through tool-augmented training, another line of research explores training-free approaches that rely on heuristic exploration to progressively locate task-relevant regions in high-resolution images~\cite{vicrop,shen2024zoomeye,dyfo}. Within this paradigm, the core distinction lies in how visual input is effectively explored. ZoomEye~\cite{shen2024zoomeye} adopts uniform spatial partitioning (Fig.~\ref{hee_fig1}(a)) and recursively selects sub-regions with higher vision–language relevance for deeper exploration, but such grid-based division often fragments semantically coherent entities and scatters critical visual cues. RAP~\cite{wang2025retrieval} processes the image differently. Specifically, it employs fixed-scale (e.g., 224/336 pixels) patch partitioning followed by scoring and top-$k$ selection over the entire image (Fig.~\ref{hee_fig1}(b)), and then reconstructs query-relevant crops according to their spatial positions into a coherent view for reasoning. Although these methods differ in search strategies, with ZoomEye performing recursive tree search with backtracking and RAP performing one-shot multi-crop retrieval with spatial layout reconstruction, they share the same fundamental limitation: the partitioning boundaries are purely geometric and entirely agnostic to object boundaries. This leads to two problems: (1)~\textit{Entity fragmentation}, where objects straddling partition boundaries are split into incomplete pieces, leaving no single candidate that contains the full target information (as also evidenced by the case study in~\cite{wang2025retrieval}); and (2)~\textit{Background interference}, where each geometrically defined region inevitably includes task-irrelevant background content. In both cases, the core operation is to crop a sub-region and resize it as input to the model.

To quantify the impact of background interference, we conduct a hierarchical decoupling analysis (Fig.~\ref{hee_fig1}(d)). A natural hypothesis is that cropping works by resizing the target region to a more suitable input resolution, enabling the model to perceive finer details. However, our background masking experiment challenges this assumption: without any cropping, simply removing background pixels progressively yields a monotonic accuracy increase across all models (Fig.~\ref{hee_fig1}(d, top); e.g., Qwen2.5-VL-7B: 56\%$\rightarrow$63\%; InternVL2.5-8B: 65\%$\rightarrow$70\%). The zoom-in experiment further corroborates this finding (Fig.~\ref{hee_fig1}(d, bottom)): contracting the view toward the GT area improves accuracy, while expanding the canvas with zero-semantic white pixels degrades performance. Together, these results reveal a key insight: the primary gain of cropping comes from background removal, not target refinement. This directly explains the bottleneck of geometry-based partitioning: regardless of how sophisticated the search strategy is, as long as the partitioning is agnostic to object boundaries, the selected regions inevitably contain background content that harms reasoning. Our conclusions are consistent with \cite{liu2025hide}, further validating the necessity of entity-aware visual perception.

Guided by these findings, we propose Hierarchical Entity Exploration (HEE), a training-free and model-agnostic framework for high-resolution visual perception. Unlike geometry-driven partitioning, HEE adopts entity-driven semantic decomposition (Fig.~\ref{hee_fig1}(c)): a frozen object detector discovers object-level entities at each level, so that the search space is organized around complete visual objects rather than being sliced by arbitrary geometric boundaries. This directly addresses both problems: detector-produced bounding boxes are naturally centered on complete objects, avoiding entity fragmentation while achieving a higher foreground ratio than geometric patches. Specifically, HEE first evaluates each visual region via a dual scoring mechanism to determine whether sufficient evidence exists to answer the question. If not, HEE selects the highest-scored region and applies a frozen object detector to extract object-level entities, which are then clustered and merged into semantically coherent sub-nodes forming the next-level search space. This process repeats recursively, with the foreground ratio progressively increasing at each level. When an exploration path fails to yield a confident answer, a backtracking mechanism returns to higher levels to explore new regions, preventing irrecoverable misses due to single-path errors.

HEE achieves significant performance gains across multiple high-resolution benchmarks and model families. On Visual Probe, it improves Qwen2.5-VL-7B from 39.1\% to 67.4\% on the Easy split and from 23.9\% to 50.0\% on the Hard split; on HR-Bench, it lifts overall accuracy from 66.5\% to 73.9\% on 4K and from 62.1\% to 71.1\% on 8K. On the MME-RealWorld benchmark, HEE brings consistent improvements across five diverse tasks (e.g., +7.5\% on Monitoring, +7.2\% on Remote Sensing), confirming its robustness in real-world visual conditions. Compared with existing training-free methods such as RAP and ZoomEye, HEE consistently achieves higher accuracy and less computational overhead. These results highlight the potential of HEE as a general, and practical solution for high-resolution visual perception. Our contributions can be summarized as:
\begin{itemize}
[leftmargin=*,itemsep=2pt,topsep=0pt,parsep=0pt,label=$\bullet$]
    \item We identify via controlled decoupling experiments that background interference is an important factor limiting geometry-based partitioning. Motivated by this, we propose Hierarchical Entity Exploration (HEE), a training-free and model-agnostic framework that replaces geometric partitioning with entity-driven semantic decomposition.
    \item HEE introduces entity-based node partitioning, a dual scoring mechanism, and confidence-guided backtracking to enable dynamic, fine-grained visual exploration organized around complete visual objects rather than arbitrary geometric regions.
    \item HEE consistently outperforms strong baselines and state-of-the-art training-free methods (i.e., RAP, ZoomEye) across multiple open source MLLMs (e.g., Qwen2.5-VL, LLaVA-OneVision) and diverse high-resolution benchmarks (Visual Probe, HR-Bench, and MME-RealWorld).
\end{itemize}

\section{Related Work}
\label{sec:related_work}

\noindent
Modern MLLMs typically consist of a pretrained visual encoder~\cite{dosovitskiy2021an,radford2021learning,jing2021meta,liu2023visual,khan2022transformers,caron2021emerging}, a frozen language model~\cite{touvron2023llama,touvron2023llama2,chu2024qwen2,bai2023qwen,yang2025qwen3,zeng2022glm}, and a lightweight multimodal connector~\cite{wang2024can,rao2022does,wang2023using,li2021align,li2023blip,cherti2023reproducible,oquab2023dinov2}. To match the resolution used in pretraining (e.g., $336\times336$ in LLaVA~\cite{liu2023visual}), input images are usually resized, which can blur or distort fine details in HR settings. Existing solutions to this challenge mainly fall into three categories: 1) cropping-based, 2) encoder-based, and 3) search-based methods.

\noindent\textbf{Cropping-Based.}
These approaches~\cite{liu2024llavanext,li2024llava,liu2024infimm} divide HR images into multiple crops, which are encoded independently via ViT~\cite{dosovitskiy2021an} and fed into the LLM. While easy to implement, they lack global coherence and incur high inference costs as resolution grows.

\noindent\textbf{Encoder-Based.}
Another line improves HR perception by upgrading the visual backbone. Works like LLaVA-HR~\cite{luo2024feast}, ConvLLaVA~\cite{ge2024convllava}, HD-SAM~\cite{ke2023segment} and MiniGemini-HD~\cite{li2024mini} adopt ConvNeXt~\cite{liu2022convnet} with higher input sizes or attention refinements. Deepseek-VL~\cite{lu2024deepseekvl} and Vary~\cite{wei2023vary} further incorporate segmentation priors such as SAM~\cite{kirillov2023segment}. These methods improve resolution handling but typically require retraining or architecture changes.

\noindent\textbf{Search-Based Methods.}
To efficiently process high-resolution images, some methods adopt a top-down search paradigm that iteratively narrows the visual scope based on task relevance.
\textit{DC$^2$}~\cite{wang2025divide} maintains a visual memory of objects and locations to retrieve relevant crops and reduce detail loss.
ZoomEye~\cite{shen2024zoomeye} applies recursive zoom-in over uniformly partitioned grids to locate informative subregions.
SEAL~\cite{wu2024v} integrates reasoning and retrieval in a unified structure to capture essential visual content.
RAP~\cite{wang2025retrieval} employs fixed-scale patch partitioning followed by scoring and top-$k$ selection, then reconstructs a compact view for inference. Unlike these methods, our approach performs entity-driven, multi-level exploration with relevance scoring and confidence-guided backtracking, enabling more reliable high-resolution perception.

\section{Method}
\label{sec:method}

\begin{figure*}[t]
    \centering
    \includegraphics[scale=0.4]{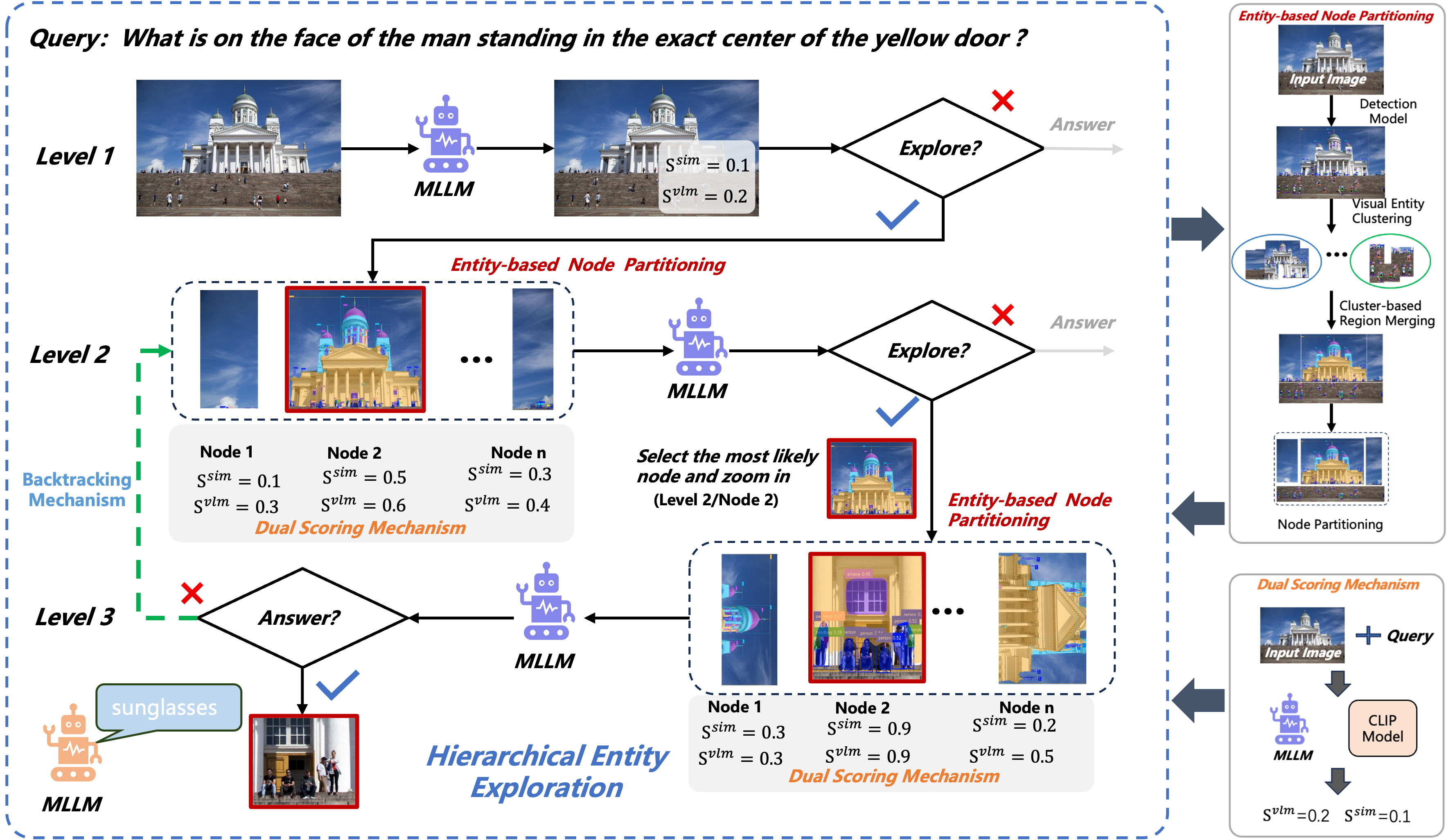}

    \caption{Overview of the HEE framework. HEE consists of four key components: (1) a dual scoring mechanism that assesses evidence sufficiency by integrating semantic similarity and model confidence; (2) entity-based node partitioning, where detected entities are clustered and merged into semantically coherent sub-regions; (3) a hierarchical exploration strategy that recursively expands the most relevant region; and (4) confidence-guided backtracking, which revisits new nodes when the active branch fails. }

    \label{hee_fig2}
    
\end{figure*}

As illustrated in Fig.~\ref{hee_fig2}, Hierarchical Entity Exploration (HEE) is a training-free and model-agnostic framework for high-resolution visual perception. Unlike previous approaches based on uniformly divided regions, HEE formulates perception as an entity-centric hierarchical tree search. At each level, the model first evaluates whether the current visual scope contains sufficient information to answer the question using a dual scoring mechanism that combines semantic similarity and model confidence. If not, the node with the highest score is selected for deeper exploration.
For the selected node (or the initial global image), HEE invokes a frozen object detector to extract object-level semantic entities. These entities are then clustered, merged into coherent regions, and partitioned into semantically consistent sub-nodes, which form the next-level search space. This recursive process proceeds adaptively until the model reaches the finest level, where individual entities are directly evaluated. If the current exploration path fails to yield a confident prediction, a backtracking mechanism restores the previous level for alternative exploration, enabling stable, human-like progressive perception without retraining or architectural modification.

\subsection{Entity-Based Node Partitioning}

After obtaining the selected node or the initial global image, HEE aims to decompose the current visual region into a set of semantically coherent and spatially complete sub-nodes, thereby constructing the search space for the next level. Given a region $\mathcal{I}^{(l)}$ at level $l$, a frozen detector $\mathcal{D}$ (implemented with DINO-X ~\cite{ren2024dino}) is applied to extract object-level semantic entities:
\begin{equation}
\mathcal{E}^{(l)} = \mathcal{D}(\mathcal{I}^{(l)}) = \{ e_i^{(l)} \}_{i=1}^{N_l},
\end{equation}
where each entity $e_i^{(l)}$ is represented by its bounding box, category label, and semantic embedding. These detected entities serve as the fundamental visual units for subsequent semantic aggregation and region partitioning. To generate a structured set of candidate nodes for the next-level search, HEE performs K-Means clustering in the semantic embedding space:
\begin{equation}
\mathcal{C}^{(l)} = \text{KMeans}(\mathcal{E}^{(l)}) = \{ C_k^{(l)} \}_{k=1}^{K},
\end{equation}
where each cluster $C_k^{(l)}$ groups semantically similar or spatially adjacent entities into a coherent conceptual region. For each cluster, the bounding boxes of its member entities are merged into a tight enclosing region:
\begin{equation}
r_k^{(l)} = \text{Merge}(C_k^{(l)}),
\end{equation}
resulting in a set of explicitly detected semantic areas denoted as $\mathcal{R}_{\text{entity}}^{(l+1)} = \{ r_k^{(l)} \}_{k=1}^{K}$.
To ensure complete spatial coverage of the scene, HEE also includes the portions of the image that are not covered by any detected entity as independent residual regions:
\begin{equation}
\mathcal{R}_{\text{residual}}^{(l+1)} = \text{MaskDiff}\left(\mathcal{I}^{(l)}, \bigcup_{k=1}^{K} r_k^{(l)} \right),
\end{equation}
where $\text{MaskDiff}(\cdot)$ denotes the pixel-wise difference between the current region mask and the union of all detected entity regions.
Finally, the next-level node set is obtained by combining the entity-based and residual regions:
\begin{equation}
\mathcal{R}^{(l+1)} = \mathcal{R}_{\text{entity}}^{(l+1)} \cup \mathcal{R}_{\text{residual}}^{(l+1)}.
\end{equation}

Each node in $\mathcal{R}^{(l+1)}$ preserves both visual and semantic features and acts as a potential branch in the hierarchical exploration tree. Through this process, HEE maintains semantic consistency while ensuring full spatial coverage, transforming unstructured visual inputs into an organized set of semantic nodes. The resulting node set $\mathcal{R}^{(l+1)}$ provides the structural foundation for the subsequent dual scoring stage, which evaluates task relevance and determines whether further recursive exploration is required.

\subsection{Dual Scoring Mechanism}

To assess the relevance between the current visual region and the question, HEE adopts a dual scoring mechanism that jointly considers \textit{semantic similarity} and \textit{task relevance}. This mechanism combines the semantic alignment capability of a frozen vision-language matching model (e.g., CLIP) with the reasoning confidence of the target VLM, providing a complementary measure of visual-question consistency.

For each candidate region $r$ (a cluster region $C_k^{(l)}$ or an entity $e_i^{(l)}$ at the final level), two scores are computed:
\begin{equation}
s^{\text{sim}}(r) = S(q, r), \quad 
s^{\text{vlm}}(r) = M(r, q),
\end{equation}
where $S(q, r)$ denotes the semantic similarity between the question $q$ and region $r$, and $M(r, q)$ is the VLM's confidence that $r$ provides sufficient evidence to answer $q$. The final score is the weighted sum of the two:
\begin{equation}
S(r) = \lambda\, s^{\text{vlm}}(r) + (1-\lambda)\, s^{\text{sim}}(r).
\end{equation}
The region with the highest score:
\begin{equation}
r^* = \arg\max_r S(r),
\end{equation}
is selected for further exploration, while a high $S(r)$ value at the current level indicates that the question can already be answered without deeper search. If all regions fall below a confidence threshold $\tau$, HEE initiates backtracking to re-examine alternative candidates from the previous level.

\subsection{Adaptive Recursive Exploration}

Building upon the dual scoring results, HEE performs adaptive recursive exploration to progressively refine its visual focus. Given the scored region set $\mathcal{R}^{(l+1)}$, the model first determines whether the current level provides sufficient confidence to answer the question. If the highest score $\max_r S(r)$ exceeds the threshold $\tau$, the exploration terminates, and the corresponding region is used for answer generation. Otherwise, HEE selects the most relevant region $r^* = \arg\max_r S(r)$ and applies entity-based partitioning to this region for deeper exploration.

At each level, all candidate regions are scored to preserve a global ranking of task relevance, but only the top-ranked region is expanded into finer sub-nodes, while other regions remain as alternative candidates for potential backtracking. This process continues until one of the following stopping conditions is met: (i) the confidence score surpasses $\tau=0.5$; (ii) the number of detected entities falls below the minimum threshold $\eta$ (e.g., only one or two entities remain); or (iii) the maximum recursion depth is reached.

If none of these conditions are met and the current region still lacks sufficient evidence, the system triggers a backtracking step that returns to the previous level and re-evaluates sub-optimal nodes. Through this adaptive search and verification loop, HEE dynamically balances depth and precision. This allows the model to zoom from coarse regions to fine-grained entities while maintaining semantic completeness and stable decision consistency across levels.

\subsection{Confidence-Guided Backtracking}

When the recursive exploration reaches the final layer and no candidate entity achieves sufficient confidence, HEE activates a confidence-guided backtracking procedure to revisit alternative search paths. This mechanism prevents the model from committing to unreliable local decisions and ensures stable reasoning under uncertain visual conditions.

Specifically, after the final-level scoring, if all entities $\{e_i\}$ yield scores below the confidence threshold $\tau$, i.e.,
\begin{equation}
\max_i S(e_i) < \tau,
\end{equation}
the system returns to the previous level and re-examines the next-best candidate region (e.g., Top-2, Top-3, etc.) based on the stored global ranking from that level. The selected region is then expanded again using entity-based partitioning and re-evaluated through the dual scoring mechanism. This backtracking process continues iteratively until a valid region surpasses the confidence threshold or all candidates are exhausted. This backtracking mechanism introduces an adaptive verification loop into the hierarchical search, allowing HEE to recover from low-confidence paths and maintain semantic consistency across levels.

\section{Experiments}
\label{sec:exp}

\subsection{Evaluation Setup}
We evaluate HEE on two high-resolution benchmarks. HR-Bench~\cite{wang2025divide} includes 8K-resolution images for fine-grained perception, covering single-instance (FSP) and cross-instance (FCP) tasks, with a cropped 4K variant also provided. To further assess robustness, we use the Visual Probe Dataset (VisualProbe)~\cite{lai2025mini}, a challenging visual search benchmark containing 4,000 training and 500 testing samples. VisualProbe features 4K-resolution images with small targets, dense distractors, and trial-and-error queries, and is divided into \textit{easy}, \textit{medium}, and \textit{hard} subsets to reflect varying levels of semantic complexity and visual ambiguity. To evaluate real-world utility, we include MME-RealWorld~\cite{zhang2024mme}, a large-scale benchmark with 13,366 high-resolution images and 29,429 human-verified QA pairs across 43 tasks. It poses substantial challenges with expert-annotated questions, diverse domains (e.g., OCR, remote sensing), and image resolutions up to 5000×5000.

HEE is implemented on top of three recent open-source vision-language models: Qwen2.5-VL-7B~\cite{bai2025qwen2}, InternVL2.5-8B~\cite{chen2024expanding}, and LLaVA-OneVision-7B~\cite{li2024llava}, without any retraining or architectural modifications. These models serve as our base systems to validate the effectiveness and generality of our HEE.

\subsection{Implementation Details}
All experiments are conducted on an NVIDIA H20 GPU with 80GB memory. We set the number of entity clusters to 4 and fix the confidence threshold $\tau$ to 0.5. The maximum number of exploration steps $T_{\text{max}}$ is set to 50 unless otherwise noted. We use DINO-X~\cite{ren2024dino} as the default object detector and adopt SigLIP for computing vision-text similarity. To ensure fair comparison with RAP, we use the same textual prompts, and set the model temperature as 0 to eliminate randomness during generation. More details are provided in the Appendix.

\begin{table*}[t]
\centering
\caption{
Comparison of \textbf{HEE} applied to several advanced MLLMs against existing methods on Visual Probe, HR-Bench~4K, and HR-Bench~8K. 
}

\renewcommand{\arraystretch}{1} 
\setlength{\tabcolsep}{6.5pt}       

\resizebox{\linewidth}{!}{
\begin{tabular}{lccccccccc}
\toprule
\multirow{2}{*}{Method} 
& \multicolumn{3}{c}{Visual Probe~\cite{lai2025mini}} 
& \multicolumn{3}{c}{HR-Bench 4K~\cite{wang2025divide}} 
& \multicolumn{3}{c}{HR-Bench 8K~\cite{wang2025divide}} \\
\cmidrule(lr){2-4} 
\cmidrule(lr){5-7} 
\cmidrule(lr){8-10}
& Easy & Medium & Hard & FSP & FCP & Overall & FSP & FCP & Overall \\
\midrule

\multicolumn{10}{c}{\textit{Open-source MLLMs}} \\
\midrule
LLaVA-v1.6-7B~\cite{liu2024llavanext} 
& -- & -- & -- 
& 49.0 & 46.8 & 47.9 
& 37.3 & 44.3 & 40.8 \\

LLaVA-HR-X-13B~\cite{luo2024feast} 
& -- & -- & -- 
& 61.3 & 46.0 & 53.6 
& 49.5 & 44.3 & 46.9 \\

LLaVA-HR-X-7B~\cite{luo2024feast} 
& -- & -- & -- 
& 57.8 & 46.3 & 52.0 
& 42.0 & 41.3 & 41.6 \\

Yi-VL-34B~\cite{young2024yi} 
& -- & -- & -- 
& 46.0 & 42.8 & 44.4 
& 39.5 & 38.5 & 39.0 \\
\midrule

\multicolumn{10}{c}{\textit{Closed-source MLLMs}} \\
\midrule

GPT-4o~\cite{hurst2024gpt} 
& 47.5 & 11.2 & 15.4 
& 70.0 & 48.0 & 59.0 
& 62.0 & 49.0 & 55.5 \\

Qwen-VL-Max~\cite{Qwen-VL} 
& 17.7 & 6.7 & 2.8 
& 65.0 & 52.0 & 58.5 
& 54.0 & 51.0 & 52.5 \\
\midrule

\multicolumn{10}{c}{\textit{Baseline and HEE}} \\
\midrule

Qwen2.5-VL-7B~\cite{bai2025qwen2} 
& 39.1 & 26.0 & 23.9 
& 81.5 & 51.5 & 66.5 
& 73.3 & 51.0 & 62.1 \\

\textit{\textbf{-w/ HEE}} 
& \textbf{67.4} & \textbf{47.0} & \textbf{50.0} 
& \textbf{92.3} & \textbf{55.5} & \textbf{73.9} 
& \textbf{88.5} & \textbf{53.8} & \textbf{71.1} \\

\rowcolor[HTML]{F5F5F5}
$\Delta(\uparrow)$ 
& \textcolor{green!60!black}{+28.3}
& \textcolor{green!60!black}{+21.0} 
& \textcolor{green!60!black}{+26.1} 
& \textcolor{green!60!black}{+10.8} 
& \textcolor{green!60!black}{+4.0} 
& \textcolor{green!60!black}{+7.4} 
& \textcolor{green!60!black}{+15.2} 
& \textcolor{green!60!black}{+2.8} 
& \textcolor{green!60!black}{+9.0} \\

InternVL2.5-8B~\cite{chen2024expanding} 
& 49.6 & 19.4 & 17.0 
& 76.8 & 56.5 & 66.6 
& 59.5 & 50.0 & 54.8 \\

\textit{\textbf{-w/ HEE}} 
& \textbf{60.3} & \textbf{42.2} & \textbf{41.5} 
& \textbf{85.0} & \textbf{59.8} & \textbf{72.4} 
& \textbf{81.8} & \textbf{51.3} & \textbf{66.5} \\

\rowcolor[HTML]{F5F5F5}
$\Delta(\uparrow)$ 
& \textcolor{green!60!black}{+10.7}
& \textcolor{green!60!black}{+22.8} 
& \textcolor{green!60!black}{+24.5} 
& \textcolor{green!60!black}{+8.2} 
& \textcolor{green!60!black}{+3.3} 
& \textcolor{green!60!black}{+5.8} 
& \textcolor{green!60!black}{+22.3} 
& \textcolor{green!60!black}{+1.3} 
& \textcolor{green!60!black}{+11.7} \\

LLaVA-ov-7B~\cite{li2024llava} 
& 36.2 & 12.5 & 13.4 
& 75.3 & 54.0 & 64.6 
& 66.5 & 48.0 & 57.3 \\

\textit{\textbf{-w/ HEE}} 
& \textbf{61.0} & \textbf{36.9} & \textbf{38.6} 
& \textbf{88.0} & \textbf{55.0} & \textbf{71.5} 
& \textbf{83.5} & \textbf{52.3} & \textbf{67.9} \\

\rowcolor[HTML]{F5F5F5}
$\Delta(\uparrow)$ 
& \textcolor{green!60!black}{+24.8}
& \textcolor{green!60!black}{+24.4} 
& \textcolor{green!60!black}{+25.2} 
& \textcolor{green!60!black}{+12.7} 
& \textcolor{green!60!black}{+1.0} 
& \textcolor{green!60!black}{+6.9} 
& \textcolor{green!60!black}{+17.0} 
& \textcolor{green!60!black}{+4.3} 
& \textcolor{green!60!black}{+10.6} \\
\bottomrule
\end{tabular}
}

\label{table:hr_bench}
\end{table*}

\begin{table*}[t]
\centering
\caption{
Performance comparison on the MME-RealWorld~\cite{zhang2024mme} across five task categories: OCR, Remote Sensing, Diagram/Table Understanding, Monitoring, and Autonomous Driving. HEE consistently improves Qwen2.5-VL-7B across all dimensions.
}
\renewcommand{\arraystretch}{1} 
\setlength{\tabcolsep}{7pt}       
\resizebox{\textwidth}{!}{%
\begin{tabular}{lcccccc}
\toprule
Method 
& OCR 
& Remote Sensing 
& Diagram/Table 
& Monitoring 
& Autonomous Driving 
& Avg. \\
\midrule
Qwen2.5-VL-7B~\cite{bai2025qwen2} 
& 84.60 & 39.22 & 77.97 & 38.52 & 28.83 & 59.99 \\
\textbf{\textit{-w/ HEE}} 
& \textbf{87.98} & \textbf{46.39} & \textbf{79.81} & \textbf{46.02} & \textbf{31.42} & \textbf{63.38} \\
\rowcolor[HTML]{F5F5F5}
$\Delta(\uparrow)$ 
& \textcolor{green!60!black}{+3.38} 
& \textcolor{green!60!black}{+7.17} 
& \textcolor{green!60!black}{+1.84} 
& \textcolor{green!60!black}{+7.50} 
& \textcolor{green!60!black}{+2.59}
& \textcolor{green!60!black}{+3.39} \\
\bottomrule
\end{tabular}%
}
\vspace{-10pt}   
\label{mme}
\end{table*}

\begin{table*}[t]
\centering
\caption{
Comparison of RAP, ZoomEye, and HEE on Visual Probe (Easy, Medium, Hard), 
HR-Bench 4K, and HR-Bench 8K. Qwen refers to the Qwen2.5-VL-7B model.
LLaVA refers to the LLaVA-OneVision-7B model. 
}
\renewcommand{\arraystretch}{1}
\setlength{\tabcolsep}{6.5pt}

\resizebox{\linewidth}{!}{
\begin{tabular}{
>{\centering\arraybackslash}m{1cm}  
ccccccccccc}
\toprule
\multirow{2}{*}{Model} &
\multirow{2}{*}{Method} &
\multicolumn{3}{c}{Visual Probe \cite{lai2025mini}} &
\multicolumn{3}{c}{HR-Bench 4K \cite{wang2025divide}} &
\multicolumn{3}{c}{HR-Bench 8K \cite{wang2025divide}} \\
\cmidrule(lr){3-5} \cmidrule(lr){6-8} \cmidrule(lr){9-11}
& & Easy & Medium & Hard & FSP & FCP & Overall & FSP & FCP & Overall \\
\midrule

\multirow{3}{*}{\parbox[c]{1.4cm}{\centering\rotatebox{90}{Qwen}}}
& RAP \cite{wang2025retrieval} 
& 52.5 & 31.7 & 34.0 
& 90.8  & 51.5 & 71.1 
& 85.5  &  53.3 & 69.4  \\

& ZoomEye \cite{shen2024zoomeye} 
& 55.3 & 35.5 & 41.5
& 87.8  &  55.0 & 71.4 
& 83.5  & 52.3  &  67.9 \\

& \textbf{HEE (Ours)} 
& \textbf{67.4} & \textbf{47.0} & \textbf{50.0} 
& \textbf{92.3} & \textbf{55.5} &\textbf{73.9}
& \textbf{88.5} & \textbf{53.8}  & \textbf{71.1}  \\
\midrule

\multirow{3}{*}{\parbox[c]{1.4cm}{\centering\rotatebox{90}{LLaVA}}}
& RAP \cite{wang2025retrieval} 
& 56.7 & 32.1 & 31.1
&84.5  & 45.3 &  64.9
&82.5   & 38.0  &  60.3 \\

& ZoomEye \cite{shen2024zoomeye} 
& 61.0 & 31.3 & 31.1
&  84.5 & 54.8  & 69.6
&  82.3 & 51.2  & 66.8 \\

& \textbf{HEE (Ours)} 
& \textbf{61.0} & \textbf{36.9} & \textbf{38.6}
& \textbf{88.0} & \textbf{55.0} & \textbf{71.5}
&  \textbf{83.5} &  \textbf{52.3} &  \textbf{67.9} \\
\bottomrule
\end{tabular}
}

\vspace{-10pt}   
\label{table:hee_comparison}

\end{table*}

\subsection{Main Results}

\noindent\textbf{Generalization across Model Families.}
Table~\ref{table:hr_bench} presents the high-resolution perception results across three representative MLLM families.
On the Visual Probe benchmark, HEE brings consistent improvements across all model families:
Qwen2.5-VL-7B gains +28.3\%, +21.0\%, and +26.1\% on the Easy, Medium, and Hard subsets;
InternVL2.5-8B gains +10.7\%, +22.8\%, and +24.5\%;
LLaVA-OneVision-7B gains +24.8\%, +24.4\%, and +25.2\%.
These results show that the entity-centric hierarchical exploration in HEE effectively enhances fine-grained perception across diverse model families by progressively focusing on task-relevant regions, capturing subtle visual cues in complex high-resolution scenes, and confirming its architecture-agnostic design and generalization capability.

\noindent\textbf{Results on Different Datasets.}
Table~\ref{table:hr_bench} also reports results on two high-resolution benchmarks: HR-Bench and Visual Probe. HR-Bench includes both 4K and 8K variants with fine-grained single-instance (FSP) and cross-instance (FCP) perception. Across both settings, HEE consistently improves all base models, achieving +5.8\%–7.4\% on 4K and +9.0\%–11.7\% on 8K (e.g., Qwen2.5-VL-7B gains +7.4\% on 4K and +9.0\% on 8K). On Visual Probe, a high-resolution benchmark focusing on open-ended VQA, the gains are even larger, showing that HEE is robust across varying resolutions and perception tasks.

\noindent\textbf{Results on Real-World Scenarios.}
We further evaluate HEE on the MME-RealWorld~\cite{zhang2024mme}, which covers five real-world categories: OCR, Remote Sensing, Diagram \& Table Understanding, Monitoring, and Autonomous Driving. Table~\ref{mme} reports the results and we can find: (i) In scenarios with high demands for fine-grained perception (e.g., Remote Sensing and Monitoring), HEE brings large gains, demonstrating its ability to capture fine-grained details in complex scenes. (ii) In other categories such as Autonomous Driving, HEE still achieves improvements, showing task generalization across diverse real-world conditions.

\begin{wraptable}{r}{0.5\textwidth}
\centering
\vspace{-25pt} 
\caption{
Efficiency comparison of RAP, ZoomEye, and HEE on HR-Bench-4K using Qwen2.5-VL-7B. 
``Throughout'' denotes the number of samples processed per minute, ``Time'' denotes the overall inference time.
}
\renewcommand{\arraystretch}{1.2} 
\setlength{\tabcolsep}{11pt} 
\resizebox{0.95\linewidth}{!}{%
\begin{tabular}{lccc}
\toprule
Method & Throughout$\uparrow$ & Time$\downarrow$ & Acc. (\%) \\ 
\midrule
RAP~\cite{wang2025retrieval}     & 1.6  &  126min & 71.1    \\ 
ZoomEye~\cite{shen2024zoomeye}   & 4.3  &  46min & 71.4   \\ 
\textbf{HEE (Ours)}              &\textbf{8.3}   &  \textbf{24min} & \textbf{73.9}    \\ 
\bottomrule
\end{tabular}%
}

\label{table:efficiency}
\vspace{-20pt} 
\end{wraptable}

\noindent\textbf{Comparison with Existing Methods.} 
We also benchmark HEE against several recent high-resolution test-time methods, including RAP~\cite{wang2025retrieval}, and ZoomEye~\cite{shen2024zoomeye}. As shown in Table~\ref{table:hee_comparison}, HEE consistently achieves the highest overall accuracy across all benchmarks and model families. Compared to RAP and ZoomEye, which rely on geometric partitioning, HEE exhibits more stable cross-resolution performance, highlighting the advantage of its hierarchical, entity-driven exploration.

Furthermore, to validate the practical efficiency of our method, we compare HEE with representative state-of-the-art high-resolution methods on HR-Bench-4K using Qwen2.5-VL-7B under the same experimental setup and hardware environment. As shown in Table~\ref{table:efficiency}, HEE achieves the best accuracy (73.9\%) while requiring the least total inference time (24 minutes), outperforming RAP (71.1\%, 126 minutes) and ZoomEye (71.4\%, 46 minutes). The efficiency gain primarily comes from HEE’s hierarchical entity selection, which avoids exhaustive region expansion and focuses computation only on semantically relevant areas. This demonstrates that our structured, entity-centric exploration not only enhances visual precision but also significantly reduces redundant computation, making HEE both effective and practically deployable for high-resolution visual tasks.

\subsection{Ablation Studies}
Unless otherwise specified, we use Qwen2.5-VL-7B as the default setting for experiments on the Visual Probe, including the Easy, Medium, and Hard subsets.

\begin{wraptable}{r}{0.5\textwidth}
\centering
\vspace{-30pt} 
\caption{Ablation study of HEE on the Visual Probe benchmark. Removing any component leads to a clear drop in accuracy.}

\renewcommand{\arraystretch}{1.2} 
\setlength{\tabcolsep}{4pt} 
\resizebox{0.48\textwidth}{!}{
\begin{tabular}{lccc}
\toprule
Setting & Easy & Medium & Hard \\
\midrule
w/o Entity Exploration & 61.7 & 44.7 & 37.7 \\
w/o Backtracking       & 60.3 & 44.8 & 41.5 \\
\midrule
\textbf{HEE (Ours)}    & \textbf{67.3} & \textbf{47.0} & \textbf{50.0} \\
\bottomrule
\end{tabular}%
}

\label{table:ablation}
\vspace{-20pt} 
\end{wraptable}

\noindent\textbf{Effects of different components of HEE. } 
Table~\ref{table:ablation} reports the contribution of each core component in HEE on the Visual Probe benchmark. Removing the entity exploration module and replacing it with uniform grid partitioning (\textit{w/o Entity Exploration}) leads to performance drops of 5.6\%, 2.3\%, and 12.3\% on the Easy, Medium, and Hard subsets, respectively. This suggests that uniform partitioning fails to capture the semantic organization of the image, while the entity-centric exploration provides a more structured representation that enables the model to focus precisely on task-relevant regions. Removing the confidence-guided backtracking module (\textit{w/o Backtracking}) results in moderate decreases of 7.0\%, 2.2\%, and 5.2\%, indicating that backtracking helps prevent error accumulation and allows recovery from low-confidence exploration paths. Overall, these results demonstrate that entity exploration, and backtracking play complementary roles in structured perception, and dynamic correction, jointly contributing to the robustness of HEE in high-resolution visual perception.

\begin{wrapfigure}{l}{0.5\textwidth}
    \centering
    \vspace{-20pt} 
    \includegraphics[width=0.42\textwidth]{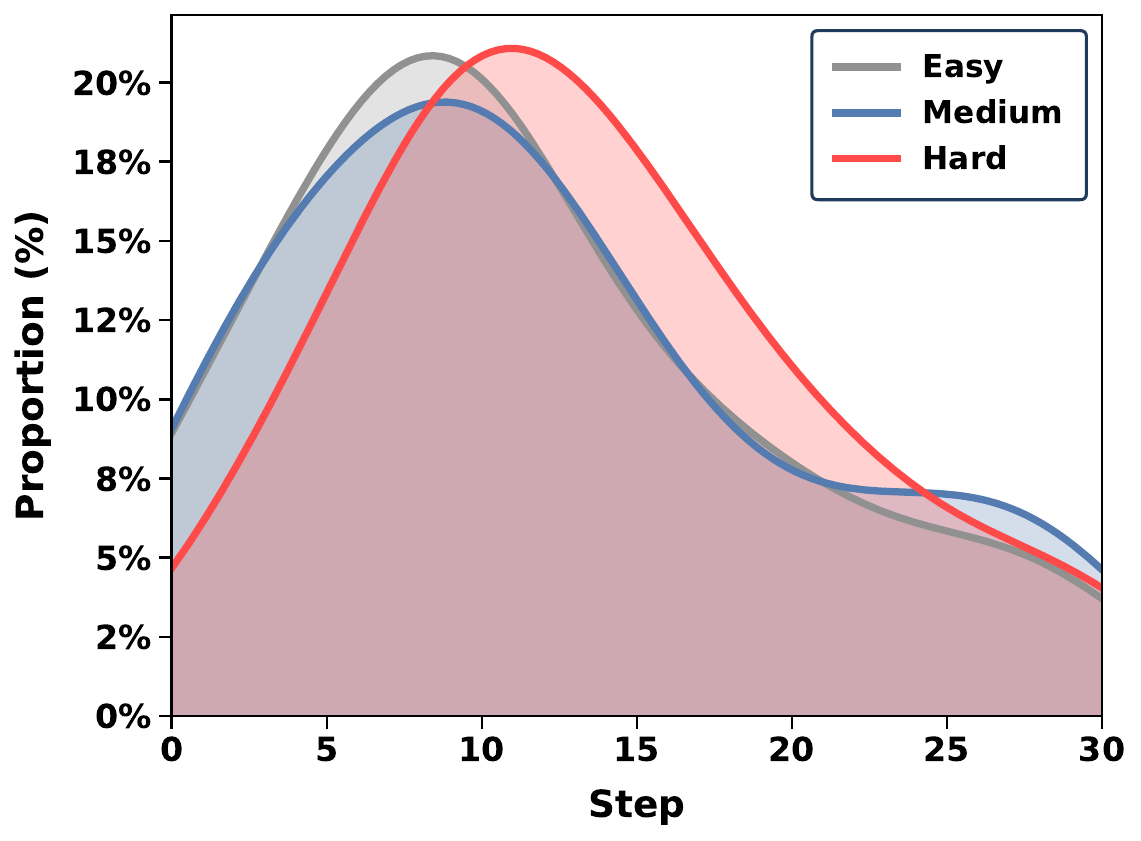}
    \vspace{-5pt}
    \caption{Distribution of interaction steps in HEE. We plot the step distribution on the Easy, Medium, and Hard splits of Visual Probe using Qwen2.5-VL-7B.}
    \label{hee_sum_fig6}
    \vspace{-20pt} 
\end{wrapfigure}

\noindent\textbf{Distribution of Interaction Steps.}
Fig.~\ref{hee_sum_fig6} reports the distribution of exploration steps required to reach the final decision across the Easy, Medium, and Hard splits of Visual Probe. All three splits exhibit a clear unimodal pattern, with the majority of queries terminating within 8--15 steps. The Easy and Medium splits display nearly identical distributions, both peaking around 10 steps, indicating that HEE identifies the correct entity clusters with minimal search depth. The Hard split shows a slightly right-shifted peak and a heavier tail, reflecting increased ambiguity and smaller visual targets in these examples. Even so, more than 60\% of hard cases converge within 15 steps. Overall, these results demonstrate that the hierarchical entity decomposition effectively suppresses irrelevant branches and avoids exhaustive search, enabling HEE to locate task-relevant regions efficiently even in high-resolution, cluttered scenes.

\begin{table}[t]
\centering
\renewcommand{\arraystretch}{1.3}
\footnotesize

\begin{minipage}[t]{0.54\textwidth}
    \centering
    \caption{Comparison of MLLM performance with different clustered-entity numbers on Visual Probe benchmark.}
    \label{table:hee_cluster}
    \vspace{2pt}
    \resizebox{\linewidth}{!}{
    \begin{tabular}{lc|ccc|c}
    \toprule
    Model & \#E. & Hard & Medium & Easy & Avg. S. \\
    \midrule
    Qwen2.5-VL-7B & -- & 23.9 & 26.0 & 39.1 & 29.7 \\
    \midrule
    \quad w/ HEE & - & 40.5 & 45.9 & 60.3 & 48.9 \\
    \quad w/ HEE & 2 & 42.5 & 45.5 & 61.7 & 49.9 \\
    \quad w/ HEE & 4 & \textbf{50.0} & \textbf{47.0} & \textbf{67.3} & \textbf{54.7} \\
    \quad w/ HEE & 8 & 37.7 & 45.1 & 62.4 & 48.4 \\
    \bottomrule
    \end{tabular}
    }
\end{minipage}
\hfill
\begin{minipage}[t]{0.44\textwidth}
    \centering
    \caption{Ablation study of the scoring formulation in Eq.~(7) on Visual Probe using Qwen2.5-VL-7B.}
    \label{table:ablation_lambda}
    \vspace{2pt}
    \resizebox{\linewidth}{!}{
    \begin{tabular}{l|ccc|c}
    \toprule
    Setting & Hard & Medium & Easy & Average \\
    \midrule
    w/o $S(q,r)$   & 43.4 & 45.9 & 64.5 & 51.3 \\
    w/o $M(r,q)$   & 21.7 & 42.9 & 56.0 & 40.2 \\
    \midrule
    $\lambda = 0.2$ & 41.5 & 44.8 & 65.9 & 50.7 \\
    $\lambda = 0.7$ & 46.2 & 46.6 & 66.7 & 53.2 \\
    \midrule
    $\lambda = 0.5$  & \textbf{50.0} & \textbf{47.0} & \textbf{67.3} & \textbf{54.7} \\
    \bottomrule
    \end{tabular}
    }
\end{minipage}
\vspace{-10pt}   
\end{table}

\noindent\textbf{Effect of Clustered-Entity Number.}
Table~\ref{table:hee_cluster} analyzes the effect of different clustered-entity numbers on Visual Probe using Qwen2.5-VL-7B. Performance improves steadily as the number of clusters increases from 2 to 4, reaching 67.3\%, 47.0\%, and 50.0\% on the Easy, Medium, and Hard subsets, respectively. This trend shows that a moderate number of entities helps the model capture diverse yet compact semantic regions, facilitating more reliable hierarchical exploration. When the number increases to 8, the performance drops notably, indicating that excessive partitioning produces fragmented representations and redundant search paths that hinder reasoning. These results reveal that the number of entities plays a role in balancing semantic diversity and exploration efficiency.

\begin{wraptable}{l}{0.5\textwidth}
\centering
\vspace{-30pt} 
\caption{
Analysis of detector coverage versus answer correctness. Results are based on Qwen2.5-VL-7B evaluated on Visual Probe.
}
\renewcommand{\arraystretch}{1.2}
\setlength{\tabcolsep}{4pt}

\resizebox{0.4\textwidth}{!}{%
\begin{tabular}{c|cc}
\toprule
\diagbox[width=7em,trim=l]{Detection}{Result}
& Correct & Wrong \\
\midrule
Detected & 63.83\% & 24.82\% \\
Not Detected & 3.55\% & 7.80\% \\
\bottomrule
\end{tabular}%
}

\label{table:detector_diagbox}
\vspace{-20pt}
\end{wraptable}

\noindent\textbf{Effect of the Dual Scoring Mechanism.}  
Table~\ref{table:ablation_lambda} studies the impact of the dual scoring function defined in Eq.~(7). Removing the semantic similarity term $S(q,r)$ or VLM’s confidence term $M(r,q)$ causes a clear performance drop, with the latter showing a stronger effect (40.2\% vs. 51.3\% on average). This suggests that both components contribute complementary cues: $S(q,r)$ captures semantic relevance between the question and visual region, while $M(r,q)$ reflects model-level confidence in visual–language alignment. Varying the balance factor $\lambda$ further reveals stable performance within a moderate range (0.2$\sim$0.7), and the best results are achieved at $\lambda=0.5$. These findings confirm that the proposed dual scoring mechanism effectively integrates semantic and confidence signals, providing a more reliable measure of task relevance during hierarchical exploration. Besides, removing SigLIP $S(q, r)$ and relying solely on the VLM's confidence still yields a 51.3 average score, suggesting that region-question relevance is primarily captured by the model itself and its confidence is reliable.

\begin{figure*}[t]
    \centering
    \includegraphics[scale=0.5]{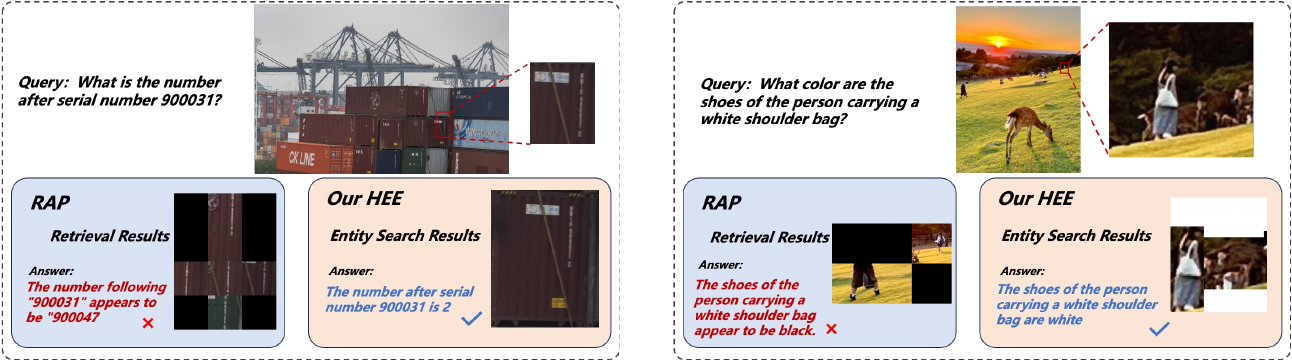}

    \caption{Comparison with RAP on fine-grained HR visual reasoning. Left: RAP retrieves irrelevant patches and predicts an incorrect serial number; HEE isolates the correct entity. Right: RAP misidentifies shoe color from background-dominant crops; HEE localizes the correct person via the shoulder-bag cue.
}
    \label{hee_sum_fig2}
\vspace{-15pt}     
\end{figure*}

\noindent\textbf{Effect of the Detection Failures.} Table~\ref{table:detector_diagbox} categorizes Qwen2.5-VL-7B's predictions on Visual Probe by whether the detector covers the answer-relevant region. Most correct predictions (63.83\%) coincide with successful detection, while 3.55\% remain correct without it, showing HEE can exploit contextual cues when localization fails. Notably, 24.82\% of errors occur despite correct detection, indicating reasoning, not detection, is the dominant bottleneck. Overall, HEE benefits from accurate detection but is not strictly dependent on it.

\begin{wraptable}{r}{0.55\textwidth}
\centering
\vspace{-25pt} 
\caption{Ablation of different detection models and CLIP encoders on Visual Probe using Qwen2.5-VL-7B.}
\renewcommand{\arraystretch}{1.22} 
\setlength{\tabcolsep}{3pt} 
\resizebox{0.53\textwidth}{!}{%
\begin{tabular}{lcccc}
\toprule
Det. Model & CLIP Model & Hard & Medium & Easy \\
\midrule
DINO-X~\cite{ren2024dino} & CLIP~\cite{radford2021learning} 
& 45.3 & 45.2 & 62.4 \\

DINO-X~\cite{ren2024dino} & SigLIP~\cite{zhai2023sigmoid} 
& \textbf{50.0} & \textbf{47.0} & \textbf{67.4} \\

YOLO-World~\cite{cheng2024yolo} & CLIP~\cite{radford2021learning} 
& 41.5 & 45.5 & 58.9 \\

YOLO-World~\cite{cheng2024yolo} & SigLIP~\cite{zhai2023sigmoid} 
& 41.5 & 47.0 & 62.4 \\
\bottomrule
\end{tabular}%
}

\label{table:ablation_detection_clip}
\vspace{-15pt} 
\end{wraptable}

\noindent\textbf{Effect of Detection and CLIP Encoders.}  
Table~\ref{table:ablation_detection_clip} examines the influence of different detection models and visual–text alignment encoders on the Visual Probe benchmark. Replacing the detector from DINO-X~\cite{ren2024dino} to YOLO-World~\cite{cheng2024yolo} leads to a consistent performance drop across all subsets, indicating that accurate entity localization is crucial for reliable hierarchical exploration. Comparing CLIP~\cite{radford2021learning} and SigLIP~\cite{zhai2023sigmoid}, the latter provides steady improvements under both detectors (e.g., +4.7\% on average with DINO-X~\cite{ren2024dino}), suggesting that stronger visual–language alignment benefits the scoring of fine-grained regions. Overall, the results highlight that high-quality detection and semantically aligned CLIP encoders are both essential to maximize HEE’s performance in high-resolution visual perception. Furthermore, when we remove SigLIP and replace DINO-X with detection outputs generated directly by the VLM itself, HEE still achieves competitive results (61.0/44.4/45.3 on Easy/Medium/Hard), showing that our HEE remains effective without any external detection or alignment modules.

\begin{figure*}[t]
    \centering
    \includegraphics[scale=0.5]{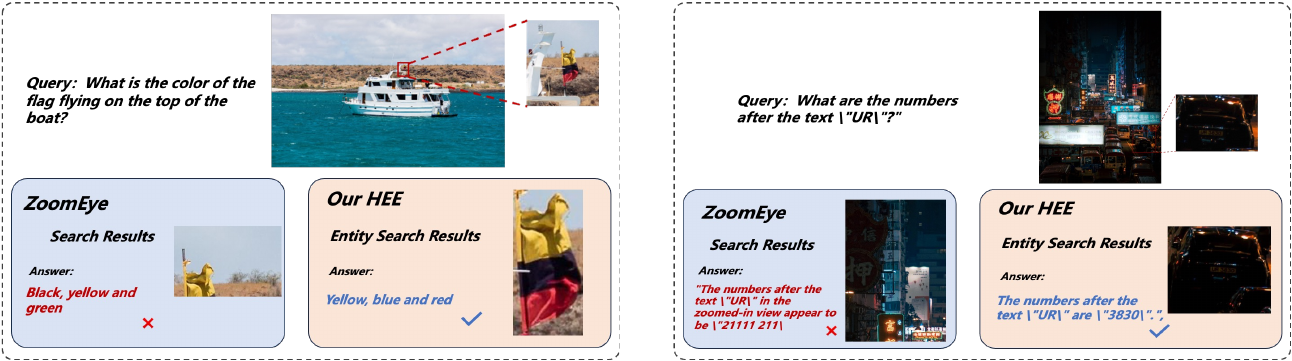}

    \caption{Comparison with ZoomEye on challenging HR scenarios. Left: ZoomEye zooms into incomplete flag regions and predicts incorrect colors; HEE isolates the full flag entity. Right: ZoomEye is distracted by bright signage and misreads the text; HEE locates the license plate entity and extracts the correct suffix.
}
    \label{hee_sum_fig3}
\vspace{-10pt}       
\end{figure*}

\begin{figure*}[t]
    \centering
    \includegraphics[scale=0.5]{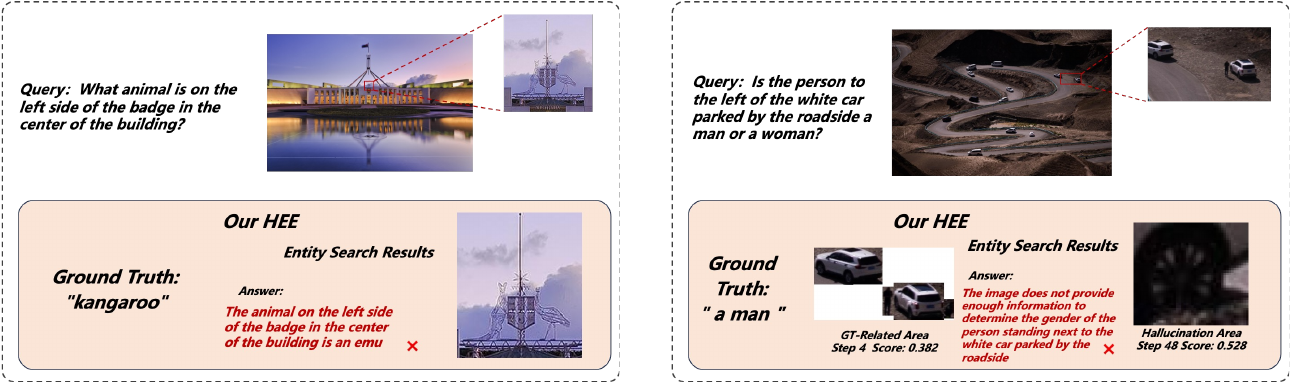}

    \caption{Failure cases of HEE. Left: HEE correctly locates the emblem region but misclassifies the kangaroo as an emu due to subtle visual similarities. Right: HEE passes through the correct region but drifts to a hallucinated high-confidence area, producing an incorrect answer.
}

    \label{hee_sum_fig5}
\vspace{-15pt}       
\end{figure*}

\noindent\textbf{Case Study.} We present qualitative comparisons in Figs.~\ref{hee_sum_fig2}--\ref{hee_sum_fig3}. Against RAP, HEE avoids retrieval noise from incomplete or irrelevant patches by progressively narrowing the search through entity-based partitioning, correctly identifying fine-grained targets (e.g., container IDs, shoe colors) that RAP misses. Against ZoomEye, HEE isolates complete semantic units (e.g., flags, license plates) rather than zooming into fragmented regions dominated by background, yielding accurate answers in cluttered scenes. We also analyze two failure modes (Fig.~\ref{hee_sum_fig5}): (i) correct localization but incorrect fine-grained recognition, where subtle visual differences (e.g., metallic sculptures) remain ambiguous even after precise decomposition, and (ii) confidence misalignment, where the model visits the correct region but drifts toward a hallucinated high-confidence area.

\section{Conclusion }
We propose Hierarchical Entity Exploration (HEE), a training-free and model-agnostic framework for high-resolution visual perception. Unlike previous methods based on geometric partitioning, HEE performs entity-centered dynamic exploration through recursive selection, dual scoring, and confidence-guided backtracking. Experiments on Visual Probe, HR-Bench, and MME-RealWorld show consistent and significant gains across diverse multimodal model families. Ablation studies confirm that entity exploration, dual scoring, and backtracking are essential for achieving fine-grained perception and stable reasoning. Moreover, HEE achieves these improvements with low computational overhead, demonstrating an efficient solution for high-resolution scenarios.


\bibliographystyle{splncs04}
\bibliography{main}

@String(CVPR= {IEEE Conf. Comput. Vis. Pattern Recog.})

@String(ICCV= {Int. Conf. Comput. Vis.})

@String(ECCV= {Eur. Conf. Comput. Vis.})

@String(ICLR = {Int. Conf. Learn. Represent.})

@String(AAAI = {AAAI})

@String(CVPR  = {CVPR})

@String(ICCV  = {ICCV})

@String(ECCV  = {ECCV})

@String(ICLR  = {ICLR})

@misc{liu2024llavanext,
    title={LLaVA-NeXT: Improved reasoning, OCR, and world knowledge},
    url={https://llava-vl.github.io/blog/2024-01-30-llava-next/},
    author={Liu, Haotian and Li, Chunyuan and Li, Yuheng and Li, Bo and Zhang, Yuanhan and Shen},
    year={2024}
}

@inproceedings{luo2024feast,
  title={Feast Your Eyes: Mixture-of-Resolution Adaptation for Multimodal Large Language Models},
  author={Luo, Gen and Zhou, Yiyi and Zhang, Yuxin and Zheng, Xiawu and Sun, Xiaoshuai and Ji, Rongrong},
  booktitle={ICLR},
  year={2025},
}

@article{wang2023cogvlm,
  title={Cogvlm: Visual expert for pretrained language models},
  author={Wang, Weihan and Lv, Qingsong and Yu, Wenmeng and Hong, Wenyi and Qi, Ji and Wang, Yan and Ji, Junhui and Yang, Zhuoyi and Zhao, Lei and XiXuan, Song and others},
  journal={NeurIPS},
  volume={37},
  pages={121475--121499},
  year={2024}
}

@article{abdin2024phi,
  title={Phi-3 technical report: A highly capable language model locally on your phone},
  author={Abdin, Marah and Jacobs, Sam Ade and Awan, Ammar Ahmad and Aneja, Jyoti and Awadallah, Ahmed and Awadalla, Hany and Bach, Nguyen and Bahree, Amit and Bakhtiari, Arash and Behl, Harkirat and others},
  journal={arXiv preprint arXiv:2404.14219},
  year={2024},
}

@article{young2024yi,
  title={Yi: Open foundation models by 01. ai},
  author={Young, Alex and Chen, Bei and Li, Chao and Huang, Chengen and Zhang, Ge and Zhang, Guanwei and Li, Heng and Zhu, Jiangcheng and Chen, Jianqun and Chang, Jing and others},
  journal={arXiv preprint arXiv:2403.04652},
  year={2024},
}

@inproceedings{liu2024improved,
  title={Improved baselines with visual instruction tuning},
  author={Liu, Haotian and Li, Chunyuan and Li, Yuheng and Lee, Yong Jae},
  booktitle={CVPR},
  pages={26296--26306},
  year={2024}
}

@inproceedings{wu2024v,
  title={V*: Guided visual search as a core mechanism in multimodal llms},
  author={Wu, Penghao and Xie, Saining},
  booktitle={CVPR},
  pages={13084--13094},
  year={2024}
}

@article{lu2024deepseekvl,
  title={Deepseek-vl: towards real-world vision-language understanding},
  author={Lu, Haoyu and Liu, Wen and Zhang, Bo and Wang, Bingxuan and Dong, Kai and Liu, Bo and Sun, Jingxiang and Ren, Tongzheng and Li, Zhuoshu and Yang, Hao and others},
  journal={arXiv preprint arXiv:2403.05525},
  year={2024}
}

@article{Qwen-VL,
  title={Qwen-VL: A Versatile Vision-Language Model for Understanding, Localization, Text Reading, and Beyond},
  author={Bai, Jinze and Bai, Shuai and Yang, Shusheng and Wang, Shijie and Tan, Sinan and Wang, Peng and Lin, Junyang and Zhou, Chang and Zhou, Jingren},
  journal={arXiv preprint arXiv:2308.12966},
  year={2023},
}

@article{ge2024convllava,
  title={ConvLLaVA: Hierarchical Backbones as Visual Encoder for Large Multimodal Models},
  author={Ge, Chunjiang and Cheng, Sijie and Wang, Ziming and Yuan, Jiale and Gao, Yuan and Song, Jun and Song, Shiji and Huang, Gao and Zheng, Bo},
  journal={arXiv preprint arXiv:2405.15738},
  year={2024},
}

@inproceedings{li2023blip,
  title={Blip-2: Bootstrapping language-image pre-training with frozen image encoders and large language models},
  author={Li, Junnan and Li, Dongxu and Savarese, Silvio and Hoi, Steven},
  booktitle={ICML},
  pages={19730--19742},
  year={2023},
}

@article{liu2024infimm,
  title={InfiMM-HD: A Leap Forward in High-Resolution Multimodal Understanding},
  author={Liu, Haogeng and You, Quanzeng and Han, Xiaotian and Wang, Yiqi and Zhai, Bohan and Liu, Yongfei and Tao, Yunzhe and Huang, Huaibo and He, Ran and Yang, Hongxia},
  journal={arXiv preprint arXiv:2403.01487},
  year={2024},
}

@inproceedings{wei2023vary,
  title={Vary: Scaling up the Vision Vocabulary for Large Vision-Language Model},
  author={Wei, Haoran and Kong, Lingyu and Chen, Jinyue and Zhao, Liang and Ge, Zheng and Yang, Jinrong and Sun, Jianjian and Han, Chunrui and Zhang, Xiangyu},
  booktitle={ECCV},
  pages={408--424},
  year={2024},
}

@inproceedings{kirillov2023segment,
  title={Segment anything},
  author={Kirillov, Alexander and Mintun, Eric and Ravi, Nikhila and Mao, Hanzi and Rolland, Chloe and Gustafson, Laura and Xiao, Tete and Whitehead, Spencer and Berg, Alexander C and Lo, Wan-Yen and others},
  booktitle={ICCV},
  pages={4015--4026},
  year={2023}
}

@article{li2024mini,
  title={Mini-Gemini: Mining the Potential of Multi-Modality Vision Language Models},
  author={Li, Yanwei and Zhang, Yuechen and Wang, Chengyao and Zhong, Zhisheng and Chen, Yixin and Chu, Ruihang and Liu, Shaoteng and Jia, Jiaya},
  journal={IEEE Transactions on Pattern Analysis and Machine Intelligence},
  volume={48},
  number={3},
  pages={3530--3543},
  year={2026},
}

@inproceedings{liu2022convnet,
  title={A convnet for the 2020s},
  author={Liu, Zhuang and Mao, Hanzi and Wu, Chao-Yuan and Feichtenhofer, Christoph and Darrell, Trevor and Xie, Saining},
  booktitle={CVPR},
  pages={11976--11986},
  year={2022}
}

@article{touvron2023llama2,
  title={Llama 2: Open foundation and fine-tuned chat models},
  author={Touvron, Hugo and Martin, Louis and Stone, Kevin and Albert, Peter and Almahairi, Amjad and Babaei, Yasmine and Bashlykov, Nikolay and Batra, Soumya and Bhargava, Prajjwal and Bhosale, Shruti and others},
  journal={arXiv preprint arXiv:2307.09288},
  year={2023},
}

@article{touvron2023llama,
  title={Llama: Open and efficient foundation language models},
  author={Touvron, Hugo and Lavril, Thibaut and Izacard, Gautier and Martinet, Xavier and Lachaux, Marie-Anne and Lacroix, Timoth{\'e}e and Rozi{\`e}re, Baptiste and Goyal, Naman and Hambro, Eric and Azhar, Faisal and others},
  journal={arXiv preprint arXiv:2302.13971},
  year={2023},
}

@article{bai2023qwen,
  title={Qwen technical report},
  author={Bai, Jinze and Bai, Shuai and Chu, Yunfei and Cui, Zeyu and Dang, Kai and Deng, Xiaodong and Fan, Yang and Ge, Wenbin and Han, Yu and Huang, Fei and others},
  journal={arXiv preprint arXiv:2309.16609},
  year={2023},
}

@inproceedings{10.1145/3664647.3681403,
author = {Wang, Wenbin and Ding, Liang and Shen, Li and Luo, Yong and Hu, Han and Tao, Dacheng},
title = {WisdoM: Improving Multimodal Sentiment Analysis by Fusing Contextual World Knowledge},
year = {2024},
isbn = {9798400706868},
booktitle={ACM MM},
pages = {2282–2291},
numpages = {10},
location = {Melbourne VIC, Australia},
}

@inproceedings{shen2024zoomeye,
  title={ZoomEye: Enhancing Multimodal LLMs with Human-Like Zooming Capabilities through Tree-Based Image Exploration},
  author={Shen, Haozhan and Zhao, Kangjia and Zhao, Tiancheng and Xu, Ruochen and Zhang, Zilun and Zhu, Mingwei and Yin, Jianwei},
  booktitle={EMNLP},
  pages={6613--6629},
year={2025}
}

@inproceedings{
dosovitskiy2021an,
title={An Image is Worth 16x16 Words: Transformers for Image Recognition at Scale},
author={Alexey Dosovitskiy and Lucas Beyer and Alexander Kolesnikov and Dirk Weissenborn and Xiaohua Zhai and Thomas Unterthiner and Mostafa Dehghani and Matthias Minderer and Georg Heigold and Sylvain Gelly and Jakob Uszkoreit and Neil Houlsby},
booktitle={ICLR},
year={2021},
}

@article{zhang2024llava,
  title={LLaVA-UHD v2: Exploiting Hierarchical Vision Granularity in MLLMs via Inverse Semantic Pyramid},
  author={Zhang, Yipeng and Liu, Yifan and Guo, Zonghao and Zhang, Yidan and Yang, Xuesong and Zhang, Xiaoying and Chen, Chi and Song, Jun and Yao, Yuan and Chua, Tat-Seng and Sun, Maosong},
  journal={AAAI},
  volume={40},
  number={15},
  pages={12934--12942},
  year={2026},
}

@article{li2024llava,
  title={LLaVA-OneVision: Easy Visual Task Transfer},
  author={Li, Bo and Zhang, Yuanhan and Guo, Dong and Zhang, Renrui and Li, Feng and Zhang, Hao and Zhang, Kaichen and Zhang, Peiyuan and Li, Yanwei and Liu, Ziwei and Li, Chunyuan},
  journal={TMLR},
  volume={2025},
  year={2025},
}

@article{hurst2024gpt,
  title={Gpt-4o system card},
  author={Hurst, Aaron and Lerer, Adam and Goucher, Adam P and Perelman, Adam and Ramesh, Aditya and Clark, Aidan and Ostrow, AJ and Welihinda, Akila and Hayes, Alan and Radford, Alec and others},
  journal={arXiv preprint arXiv:2410.21276},
  year={2024},
}

@inproceedings{
ji2026tree,
title={Tree Search for {LLM} Agent Reinforcement Learning},
author={Yuxiang Ji and Ziyu Ma and Yong Wang and Guanhua Chen and Xiangxiang Chu and Liaoni Wu},
booktitle={The Fourteenth International Conference on Learning Representations},
year={2026},
url={https://openreview.net/forum?id=ZpQwAFhU13}
}

@article{ma2024drvideo,
  title={Drvideo: Document retrieval based long video understanding},
  author={Ma, Ziyu and Gou, Chenhui and Shi, Hengcan and Sun, Bin and Li, Shutao and Rezatofighi, Hamid and Cai, Jianfei},
  journal={arXiv preprint arXiv:2406.12846},
  year={2024}
}

@article{ma2024gerea,
  title={GeReA: Question-Aware Prompt Captions for Knowledge-based Visual Question Answering},
  author={Ma, Ziyu and Li, Shutao and Sun, Bin and Cai, Jianfei and Long, Zuxiang and Ma, Fuyan},
  journal={arXiv preprint arXiv:2402.02503},
  year={2024}
}

@article{Ma_Gou_Hu_Wang_Zhuang_Cai_2026, title={Where and What Matters: Sensitivity-Aware Task Vectors for Many-Shot Multimodal In-Context Learning}, volume={40}, url={https://ojs.aaai.org/index.php/AAAI/article/view/37733}, DOI={10.1609/aaai.v40i10.37733}, abstractNote={Large Multimodal Models (LMMs) have shown promising in-context learning (ICL) capabilities, but scaling to many-shot settings remains difficult due to limited context length and high inference cost. To address these challenges, task-vector-based methods have been explored by inserting compact representations of many-shot in-context demonstrations into model activations. However, existing task-vector-based methods either overlook the importance of where to insert task vectors or struggle to determine suitable values for each location. To this end, we propose a novel Sensitivity-aware Task Vector insertion framework (STV) to figure out where and what to insert. Our key insight is that activation deltas across query-context pairs exhibit consistent structural patterns, providing a reliable cue for insertion. Based on the identified sensitive-aware locations, we construct a pre-clustered activation bank for each location by clustering the activation values, and then apply reinforcement learning to choose the most suitable one to insert. We evaluate STV across a range of multimodal models (e.g., Qwen-VL, Idefics-2) and tasks (e.g., VizWiz, OK-VQA), demonstrating its effectiveness and showing consistent improvements over previous task-vector-based methods with strong generalization.}, number={10}, journal={Proceedings of the AAAI Conference on Artificial Intelligence}, author={Ma, Ziyu and Gou, Chenhui and Hu, Yiming and Wang, Yong and Zhuang, Bohan and Cai, Jianfei}, year={2026}, month={Mar.}, pages={7892-7900} }

@article{ma2026skillclaw,
  title={SkillClaw: Let Skills Evolve Collectively with Agentic Evolver},
  author={Ma, Ziyu and Yang, Shidong and Ji, Yuxiang and Wang, Xucong and Wang, Yong and Hu, Yiming and Huang, Tongwen and Chu, Xiangxiang},
  journal={arXiv preprint arXiv:2604.08377},
  year={2026}
}

@inproceedings{jing2023deep,
  title={Deep graph reprogramming},
  author={Jing, Yongcheng and Yuan, Chongbin and Ju, Li and Yang, Yiding and Wang, Xinchao and Tao, Dacheng},
  booktitle={CVPR},
  pages={24345--24354},
  year={2023},
}

@inproceedings{jing2021meta,
  title={Meta-aggregator: Learning to aggregate for 1-bit graph neural networks},
  author={Jing, Yongcheng and Yang, Yiding and Wang, Xinchao and Song, Mingli and Tao, Dacheng},
  booktitle={ICCV},
  pages={5301--5310},
  year={2021},
}

@article{wang2024can,
  title={Can linguistic knowledge improve multimodal alignment in vision-language pretraining?},
  author={Wang, Fei and Ding, Liang and Rao, Jun and Liu, Ye and Shen, Li and Ding, Changxing},
  journal={ACM Transactions on Multimedia Computing, Communications and Applications},
  volume={20},
  number={12},
  pages={1--22},
  year={2024},
  publisher={ACM New York, NY}
}

@inproceedings{wang2023using,
  title={Using Self-Supervised Dual Constraint Contrastive Learning for Cross-Modal Retrieval.},
  author={Wang, Xintong and Li, Xiaoyu and Ding, Liang and Zhao, Sanyuan and Biemann, Chris},
  booktitle={ECAI},
  pages={2552--2559},
  year={2023}
}

@inproceedings{rao2022does,
  title={Where Does the Performance Improvement Come From? -A Reproducibility Concern about Image-Text Retrieval},
  author={Rao, Jun and Wang, Fei and Ding, Liang and Qi, Shuhan and Zhan, Yibing and Liu, Weifeng and Tao, Dacheng},
  booktitle={SIGIR},
  pages={2727--2737},
  year={2022}
}

@inproceedings{zhai2023sigmoid,
  title={Sigmoid loss for language image pre-training},
  author={Zhai, Xiaohua and Mustafa, Basil and Kolesnikov, Alexander and Beyer, Lucas},
  booktitle={ICCV},
  pages={11975--11986},
  year={2023}
}

@article{visual-cot,
  title={Visual cot: Advancing multi-modal language models with a comprehensive dataset and benchmark for chain-of-thought reasoning},
  author={Shao, Hao and Qian, Shengju and Xiao, Han and Song, Guanglu and Zong, Zhuofan and Wang, Letian and Liu, Yu and Li, Hongsheng},
  journal={NeurIPS},
  volume={37},
  pages={8612--8642},
  year={2024}
}

@inproceedings{deepeyes,
  title={DeepEyes: Incentivizing ``Thinking with Images'' via Reinforcement Learning},
  author={Zheng, Ziwei and Yang, Michael and Hong, Jack and Zhao, Chenxiao and Xu, Guohai and Yang, Le and Shen, Chao and Yu, Xing},
  booktitle={ICLR},
  year={2026},
}

@inproceedings{unvisual_cot,
  title={Unsupervised Visual Chain-of-Thought Reasoning via Preference Optimization},
  author={Zhao, Kesen and Zhu, Beier and Sun, Qianru and Zhang, Hanwang},
  booktitle={ICCV},
  pages={2303--2312},
year={2025},
}

@inproceedings{
investigating,
title={Investigating the Catastrophic Forgetting in Multimodal Large Language Model},
author={Yuexiang Zhai and Shengbang Tong and Xiao Li and Mu Cai and Qing Qu and Yong Jae Lee and Yi Ma},
booktitle={CPAL},
year={2023},
}

@inproceedings{limit-of-rlvr,
  title={Does Reinforcement Learning Really Incentivize Reasoning Capacity in LLMs Beyond the Base Model?},
  author={Yue, Yang and Chen, Zhiqi and Lu, Rui and Zhao, Andrew and Wang, Zhaokai and Yue, Yang and Song, Shiji and Huang, Gao},
  booktitle={NeurIPS},
  volume={38},
  pages={57654--57689},
year={2025},
}

@inproceedings{
vicrop,
title={MLLMs Know Where to Look: Training-free Perception of Small Visual Details with Multimodal LLMs},
author={Jiarui Zhang and Mahyar Khayatkhoei and Prateek Chhikara and Filip Ilievski},
booktitle={ICLR},
year={2025},
}

@inproceedings{dyfo,
  title={Dyfo: A training-free dynamic focus visual search for enhancing lmms in fine-grained visual understanding},
  author={Li, Geng and Xu, Jinglin and Zhao, Yunzhen and Peng, Yuxin},
  booktitle={CVPR},
  pages={9098--9108},
  year={2025}
}

@inproceedings{wang2025retrieval,
  title={Retrieval-Augmented Perception: High-resolution Image Perception Meets Visual RAG},
  author={Wang, Wenbin and Jing, Yongcheng and Ding, Liang and Wang, Yingjie and Shen, Li and Luo, Yong and Du, Bo and Tao, Dacheng},
  booktitle={ICML},
  pages={63290--63307},
  year={2025},
  volume={267},
}

@inproceedings{liu2025hide,
  title={HiDe: Rethinking The Zoom-IN method in High Resolution MLLMs via Hierarchical Decoupling},
  author={Liu, Xianjie and Hu, Yiman and Zou, Yixiong and Wu, Liang and Xu, Jian and Zheng, Bo},
  booktitle={ICML},
  year={2026},
}

@article{chen2024expanding,
  title={Expanding performance boundaries of open-source multimodal models with model, data, and test-time scaling},
  author={Chen, Zhe and Wang, Weiyun and Cao, Yue and Liu, Yangzhou and Gao, Zhangwei and Cui, Erfei and Zhu, Jinguo and Ye, Shenglong and Tian, Hao and Liu, Zhaoyang and others},
  journal={arXiv preprint arXiv:2412.05271},
  year={2024}
}

@article{bai2025qwen2,
  title={Qwen2. 5-vl technical report},
  author={Bai, Shuai and Chen, Keqin and Liu, Xuejing and Wang, Jialin and Ge, Wenbin and Song, Sibo and Dang, Kai and Wang, Peng and Wang, Shijie and Tang, Jun and others},
  journal={arXiv preprint arXiv:2502.13923},
  year={2025}
}

@article{ren2024dino,
  title={Dino-x: A unified vision model for open-world object detection and understanding},
  author={Ren, Tianhe and Chen, Yihao and Jiang, Qing and Zeng, Zhaoyang and Xiong, Yuda and Liu, Wenlong and Ma, Zhengyu and Shen, Junyi and Gao, Yuan and Jiang, Xiaoke and others},
  journal={arXiv preprint arXiv:2411.14347},
  year={2024}
}

@inproceedings{cheng2024yolo,
  title={Yolo-world: Real-time open-vocabulary object detection},
  author={Cheng, Tianheng and Song, Lin and Ge, Yixiao and Liu, Wenyu and Wang, Xinggang and Shan, Ying},
  booktitle={CVPR},
  pages={16901--16911},
  year={2024}
}

@inproceedings{radford2021learning,
  title={Learning transferable visual models from natural language supervision},
  author={Radford, Alec and Kim, Jong Wook and Hallacy, Chris and Ramesh, Aditya and Goh, Gabriel and Agarwal, Sandhini and Sastry, Girish and Askell, Amanda and Mishkin, Pamela and Clark, Jack and others},
  booktitle={ICML},
  pages={8748--8763},
  year={2021},
}

@inproceedings{lai2025mini,
  title={Mini-o3: Scaling Up Reasoning Patterns and Interaction Turns for Visual Search},
  author={Lai, Xin and Li, Junyi and Li, Wei and Liu, Tao and Li, Tianjian and Zhao, Hengshuang},
  booktitle={ICLR},
  year={2026},
}

@inproceedings{wang2025divide,
  title={Divide, conquer and combine: A training-free framework for high-resolution image perception in multimodal large language models},
  author={Wang, Wenbin and Ding, Liang and Zeng, Minyan and Zhou, Xiabin and Shen, Li and Luo, Yong and Yu, Wei and Tao, Dacheng},
  booktitle={AAAI},
  volume={39},
  pages={7907--7915},
  year={2025}
}

@inproceedings{zhang2024mme,
  title={MME-RealWorld: Could Your Multimodal LLM Challenge High-Resolution Real-World Scenarios that are Difficult for Humans?},
  author={Zhang, Yi-Fan and Zhang, Huanyu and Tian, Haochen and Fu, Chaoyou and Zhang, Shuangqing and Wu, Junfei and Li, Feng and Wang, Kun and Wen, Qingsong and Zhang, Zhang and others},
  booktitle={ICLR},
  year={2025},
}

@article{ke2023segment,
  title={Segment anything in high quality},
  author={Ke, Lei and Ye, Mingqiao and Danelljan, Martin and Tai, Yu-Wing and Tang, Chi-Keung and Yu, Fisher and others},
  journal={NeurIPS},
  volume={36},
  pages={29914--29934},
  year={2023}
}

@article{chu2024qwen2,
  title={Qwen2-audio technical report},
  author={Chu, Yunfei and Xu, Jin and Yang, Qian and Wei, Haojie and Wei, Xipin and Guo, Zhifang and Leng, Yichong and Lv, Yuanjun and He, Jinzheng and Lin, Junyang and others},
  journal={arXiv preprint arXiv:2407.10759},
  year={2024}
}

@article{yang2025qwen3,
  title={Qwen3 technical report},
  author={Yang, An and Li, Anfeng and Yang, Baosong and Zhang, Beichen and Hui, Binyuan and Zheng, Bo and Yu, Bowen and Gao, Chang and Huang, Chengen and Lv, Chenxu and others},
  journal={arXiv preprint arXiv:2505.09388},
  year={2025}
}

@article{li2021align,
  title={Align before fuse: Vision and language representation learning with momentum distillation},
  author={Li, Junnan and Selvaraju, Ramprasaath and Gotmare, Akhilesh and Joty, Shafiq and Xiong, Caiming and Hoi, Steven Chu Hong},
  journal={NeurIPS},
  volume={34},
  pages={9694--9705},
  year={2021}
}

@inproceedings{zeng2022glm,
  title={GLM-130B: An Open Bilingual Pre-trained Model},
  author={Zeng, Aohan and Liu, Xiao and Du, Zhengxiao and Wang, Zihan and Lai, Hanyu and Ding, Ming and Yang, Zhuoyi and Xu, Yifan and Zheng, Wendi and Xia, Xiao and others},
  booktitle={ICLR},
  year={2023},
}

@article{liu2023visual,
  title={Visual instruction tuning},
  author={Liu, Haotian and Li, Chunyuan and Wu, Qingyang and Lee, Yong Jae},
  journal={NeurIPS},
  volume={36},
  pages={34892--34916},
  year={2023}
}

@article{oquab2023dinov2,
  title={DINOv2: Learning Robust Visual Features without Supervision},
  author={Oquab, Maxime and Darcet, Timoth{\'e}e and Moutakanni, Th{\'e}o and Vo, Huy and Szafraniec, Marc and Khalidov, Vasil and Fernandez, Pierre and Haziza, Daniel and Massa, Francisco and El-Nouby, Alaaeldin and others},
  journal={TMLR},
  year={2024},
}

@inproceedings{cherti2023reproducible,
  title={Reproducible scaling laws for contrastive language-image learning},
  author={Cherti, Mehdi and Beaumont, Romain and Wightman, Ross and Wortsman, Mitchell and Ilharco, Gabriel and Gordon, Cade and Schuhmann, Christoph and Schmidt, Ludwig and Jitsev, Jenia},
  booktitle={CVPR},
  pages={2818--2829},
  year={2023}
}

@inproceedings{caron2021emerging,
  title={Emerging properties in self-supervised vision transformers},
  author={Caron, Mathilde and Touvron, Hugo and Misra, Ishan and J{\'e}gou, Herv{\'e} and Mairal, Julien and Bojanowski, Piotr and Joulin, Armand},
  booktitle={ICCV},
  pages={9650--9660},
  year={2021}
}

@article{khan2022transformers,
  title={Transformers in vision: A survey},
  author={Khan, Salman and Naseer, Muzammal and Hayat, Munawar and Zamir, Syed Waqas and Khan, Fahad Shahbaz and Shah, Mubarak},
  journal={ACM Computing Surveys (CSUR)},
  volume={54},
  number={10s},
  pages={1--41},
  year={2022},
  publisher={ACM New York, NY}
}
\end{document}